\pdfoutput=1

\documentclass[11pt]{article}

\usepackage[]{ACL2023}
\usepackage{multirow} 
\usepackage{array}
\usepackage{tabularx}
\usepackage{soul} 
\usepackage{color, xcolor}
\usepackage{float}
\usepackage{times}
\usepackage{latexsym}
\usepackage{amsthm}
\usepackage[T1]{fontenc}
\usepackage{graphicx}
\usepackage{float}
\usepackage{booktabs}
\usepackage{pifont} 
\usepackage[utf8]{inputenc}
\usepackage{amsmath}
\usepackage{microtype}

\usepackage{inconsolata}
\usepackage{amssymb}
\usepackage{graphicx} 
\usepackage{subcaption} 
\usepackage{array}
\usepackage{longtable}
\usepackage{blindtext}
\usepackage{cleveref}
\usepackage{enumerate}
\usepackage{array}
\usepackage{colortbl}
\usepackage{bbding}
\usepackage{tikz}
\usepackage{graphicx}
\usepackage{subcaption}
\usepackage{enumitem}
\usepackage{amsmath}
\usepackage{amsfonts}
\usepackage{bm}
\usepackage{bbold}
\usepackage{graphicx}
\usepackage{enumerate}
\usepackage{caption}
\usepackage{graphicx}
\usepackage{multirow}
\usepackage{xcolor}
\usepackage{subcaption}
\usepackage{colortbl}
\usepackage{algorithm}
\usepackage{caption}
\definecolor{goldenyellow}{rgb}{1.0, 0.93, 0.7}
\definecolor{deepblue}{rgb}{0.0, 0.25, 0.5}
\usepackage{xcolor}
\usepackage{soul}

\title{\textsc{EVER}: Mitigating Hallucination in Large Language Models through Real-Time Verification and Rectification}

\author{Haoqiang Kang$^{1,2}$, Juntong Ni$^{3}$, Huaxiu Yao$^{1}$\\
  $^{1}$UNC-Chapel Hill, $^{2}$University of Washington, $^{3}$Emory University\\
  \texttt{haoqik@uw.edu}, \texttt{huaxiu@cs.unc.edu}
}

\begin{document}

\maketitle

\begin{abstract}
Large Language Models (LLMs) have demonstrated remarkable proficiency in generating fluent text. However, they often encounter the challenge of generating inaccurate or hallucinated content. This issue is common in both non-retrieval-based generation and retrieval-augmented generation approaches, and existing post-hoc rectification methods may not address the accumulated hallucination errors that may be caused by the "snowballing" issue, especially in reasoning tasks. To tackle these challenges, we introduce a novel approach called Real-time Verification and Rectification (\textsc{\textsc{EVER}}). Instead of waiting until the end of the generation process to rectify hallucinations, \textsc{\textsc{EVER}} employs a real-time, step-wise generation and hallucination rectification strategy. Apart from directly mitigating hallucination, we further demonstrate that both the EVER-rectified response and the original one can serve as preference data to enhance the factuality of the model through preference tuning. When compared to both retrieval-based and non-retrieval-based baselines, \textsc{\textsc{EVER}} demonstrates a significant improvement in generating trustworthy and factually accurate text across a diverse range of tasks, including biography generation and multi-hop reasoning. 

\end{abstract}

\section{Introduction}
Recent years have witnessed remarkable progress in the field of Large Language Models (LLMs), which are increasingly adept at generating coherent, contextually fluent responses. Despite this, they are still prone to hallucination which is defined as the generated content is nonsensical or unfaithful to a reference content~\cite{ji2023survey, zhang2023siren}. Hallucination can be categorized into two types: intrinsic and extrinsic. Intrinsic hallucinations happen when the generated content is contradictory to the reference. Extrinsic hallucinations, meanwhile, are the content that, while seemingly plausible, cannot be verified by evidence, typically appearing as imaginative concoctions or guesses made by the model~\cite{min2023factscore, sun2023head, pmlr-v202-kandpal23a}. 

Due to the infrequent updates of an LLM's parametric knowledge base, utilizing external knowledge has shown significant leap in enhancing factuality by providing up-to-date content \cite{lewis2020retrieval}. Prior retrieval-based mitigation methods of LLM hallucination can be categorized into two categories: pre-generation, and post-generation methods. The pre-generation methods \cite{lewis2020retrieval, vu2023freshllms, asai2023self} optimize the retrieved content to be more accurate, relevant and supportive. But these methods may still produce detailed factual errors, particularly in long-form generation if there is no mechanism for post-generation checks or revisions. Another line of work focuses on enhancing the attribution of text post-generation \cite{gao2022attributed, gou2023critic, peng2023check}. However, these post-hoc editing methods do not account for the "snowballing" issue of hallucinations \cite{zhang2023language}, where initial factual errors can lead to a series of accumulated errors, and they require increasingly complex revisions to mitigate its impact.

To address these challenges, we propose the R\textbf{E}al-Time \textbf{VE}rification and \textbf{R}ectification (\textbf{\textsc{\textsc{EVER}}}) framework. Instead of mitigating hallucination until the end of generation, \textsc{\textsc{EVER}} employs real-time validation to identify both intrinsic and extrinsic hallucinations, mitigating these issues during the generation process to prevent error propagation. The process involves three stages: generation, validation, and rectification. First, a LLM generates an initial sentence based on a prompt, which may include externally retrieved knowledge, such as Retrieval-Augmented Generation (RAG)~\cite{lewis2020retrieval}. Then, it validates the correctness of each fact-related concept in the sentence by identifying intrinsic and extrinsic hallucinations. In the rectification stage, any detected errors are corrected based on the type of hallucinations identified. The rectified sentence then undergoes another round of validation. If extrinsic hallucinations persist, depending on the task, we either flag the sentence with a warning to alert users to potential issues or abstain from answering the question, which enhances the trustworthiness of the generated content. In addition to directly mitigating hallucination during the generation process, we further explore the utilization of the \textsc{EVER}-generated data to construct preference data pairs, essentially enhancing the factuality of the model through preference tuning.

Our primary contribution of this paper is \textsc{\textsc{EVER}}, which introduces a novel approach to mitigate hallucinations in LLM. Compared to the state-of-the-art prior methods, our results demonstrate the effectiveness of this approach in directly reducing hallucinations in two tasks: long-form biography generation and reasoning. Furthermore, we show the compatibility of \textsc{\textsc{EVER}}, which can serve as a complement to the traditional RAG method. Lastly, we demonstrate that \textsc{EVER}-rectified response can lead to better preference data and enhance the factuality of LLM with preference tuning.

\section{Real-Time Verification and Rectification}
\begin{figure*}[ht]
    \small
    \begin{center}
        \includegraphics[width=0.96\textwidth]{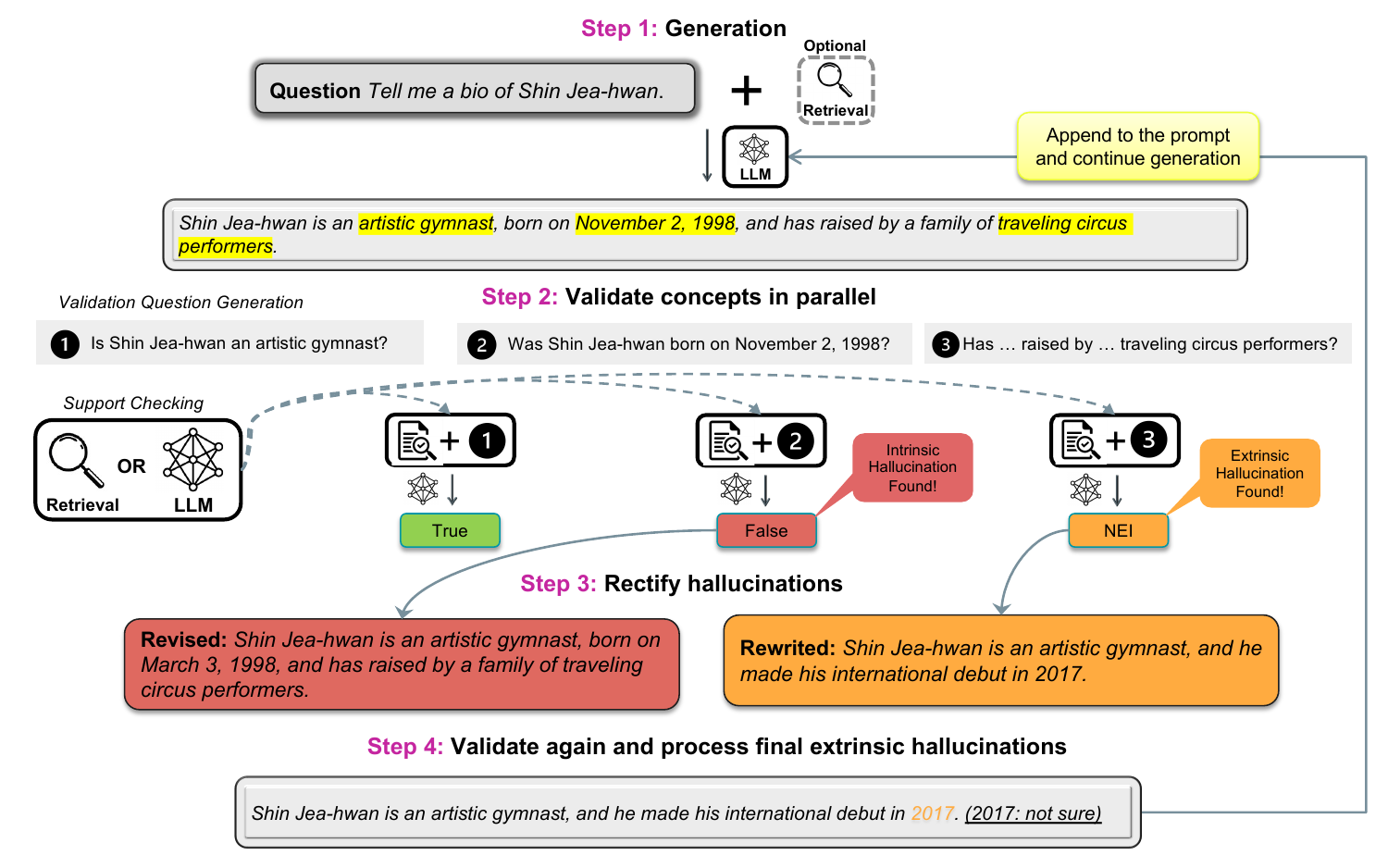}
        \vspace{-0.5em}
        \caption{Overview of \textsc{\textsc{EVER}} pipeline in the biography generation task. \textsc{\textsc{EVER}} proactively identifies and rectifies concept-level hallucinations before each new sentence generation. Also, it flags any remaining extrinsic hallucinations after a single round of rectification, thereby enhancing the trustworthiness of the output.}
        \label{fig:pipeline}
        \vspace{-1.5em}
    \end{center}

\end{figure*}

In this section, we firstly detail our method, R\textbf{E}al-time \textbf{VE}rification and \textbf{R}ectification (\textbf{\textsc{\textsc{EVER}}}), whose framework with one representative example is shown in Figure~\ref{fig:pipeline}. \textsc{\textsc{EVER}} aims to mitigate hallucinations in language model outputs by immediately validating each generated sentence during the generation period, which helps prevent error propagation. Secondly, as shown in Figure~\ref{fig:teaser}, we use \textsc{EVER}-recified response to construct better preference data to align LLM to become more factual by using preference tuning.

\subsection{Prompting-based Hallucination Verification and Mitigation}
We first present how to use EVER to directly mitigate hallucination of LLM during the response generation period.
\subsubsection{Generation}
The first stage is to generate the initial sentence given the prompt. Based on if an external knowledge is used in the prompt, we categorize the generation method to two categories: 
\vspace{-0.5em}
\begin{itemize}[leftmargin=*]
    \item \textbf{Non-retrieval Generation}: In non-retrieval generation, the LLM is provided with a query and is prompted to generate a response based solely on its internal knowledge without referring to external data sources.
    \vspace{-0.5em}
    \item \textbf{Retrieval-Augmented Generation (RAG)}: In RAG~\cite{lewis2020retrieval}, the LLM is presented with the context in the prompt.
    \vspace{-0.5em}
\end{itemize}

After determining the generation category in \textsc{\textsc{EVER}}, we adopt a real-time generation and verification strategy to mitigate the "snowballing issue" in text generation~\cite{zhang2023language, varshney2023stitch}. This effect arises when early inaccuracies or hallucinations in the text result in compounded errors in subsequent sentences. By addressing hallucinations on a real-time basis, our strategy significantly reduces the likelihood of errors propagating throughout the entire text, ensuring that early hallucinations do not have a significant impact on later generated content. Therefore, we transition to the validation and hallucination correction phases upon generating a new sentence.

\subsubsection{Concept-Level Validation}
\label{sec:validation}
In the validation stage, we meticulously evaluate the generated sentence at a concept-level, with the goal of identifying the occurrence of hallucinations and classifying them as either intrinsic or extrinsic hallucinations. The entire validation phase includes three steps: key concepts identification, validation question generation, and support checking. We detail these steps as follows:

\noindent \textbf{Key Concepts Identification.} In key concepts identification step, we leverage the in-context learning ability of the model to extracts factual-related concepts from the generated sentence. We extract all potential concepts that might cause hallucination, such as dates, numbers, jobs, locations, etc. For example, as shown in Figure \ref{fig:pipeline}, in the sentence "Shin Jea-hwan is an artistic gymnast, born on November 2, 1998, and has raised by a family of traveling circus performers.", we extract the concepts of "artistic gymnast", "November 2, 1998", and "traveling circus performers".

\noindent \textbf{Validation Question Generation.}
Once the key concepts are identified, we will use the model to generate validation questions. These validation questions are Yes/No questions constructed to verify the accuracy of the concepts in the initial sentence. For example, in Figure \ref{fig:pipeline}, for the extracted concept of "artistic gymnast", the corresponding validation question is "Is Shin Jea-hwan an artistic gymnast?"

\noindent \textbf{Support Checking.} Then, in the last step, we use few-shot Chain of Thought (CoT) prompting~\cite{wei2022chain} to guide the model to choose one of three flags for each validation question based on the evidence: \texttt{True}, \texttt{False}, or \texttt{Not Enough Information (NEI)}. A \texttt{True} flag indicates that the evidence supports the generated concept, whereas a \texttt{False} flag signifies that the generated concept is in contradiction with the evidence, pointing towards an intrinsic hallucination. The \texttt{NEI} flag is assigned when no related evidence is found, suggesting the presence of an extrinsic hallucination. To compare the effect of retrieval on our method, we test on the following two strategies.
\begin{itemize}[leftmargin=*]
\vspace{-0.5em}
    \item \textbf{Self-query}: Based on the validation question, we prompt the LLM to directly answer the question by choosing from the three labels. 
    \vspace{-0.5em}
    \item \textbf{Evidence Retrieval}: This mode leverages external knowledge source to gather evidence that can help answer the validation question.
\end{itemize}

\subsubsection{Rectifying Hallucination}

After the validation stage, if hallucination is detected, i.e., at least one validation question is assigned the flag \texttt{False} or \texttt{NEI}, \textsc{\textsc{EVER}} aims to rectify the corresponding sentence based on the evidence gathered, including two revision categories:

\noindent \textbf{Intrinsic Hallucination Revision.}
Intrinsic Hallucinations refer to instances where the generated output contradicts the source content. These hallucination will be revised based on the evidence retrieved from last step. The primary objective is to align each entity or fact with verifiable truths.

\noindent \textbf{Extrinsic Hallucination Rewrite.}
Extrinsic Hallucinations are defined as generated outputs that cannot be verified against the source content, meaning the output is neither supported nor refuted by the evidence.  When confronted with such situations, the entire sentence undergoes a rewrite, taking into account feedback that pinpoints the issue and uses the retrieved evidence as a reference.

\subsubsection{Processing the Remaining Extrinsic Hallucination}

After completing the rectification phase, the refined sentence undergoes revalidation. If intrinsic hallucinations cannot be fully rectified with a single round of rectification, we conduct additional rounds. It's important to note that, in most scenarios, one round of rectification is empirically sufficient to eliminate all intrinsic hallucinations (see detailed analysis in Appendix~\ref{app:multi-round}). In such cases, if a sentence still exhibits extrinsic hallucinations, depending on the tasks, we will further refine it. For example, in short-form generation, if there is no other verified correct answers, we will abstain from answering the question to maintain honesty. In long-form generation, we will mark it with a final warning flag, "\texttt{not sure}," indicating the presence of extrinsic hallucination and enhancing the trustworthiness of the generated content. Acknowledging limitations and errors in generated content promotes transparency and a reliable user experience. Since completely rectifying all extrinsic hallucinations can be challenging, the warning signal effectively assists users in utilizing the generated content.

\subsection{Enhancing Factuality of Model via Preference Tuning}

\begin{figure}[tp]
    \centering
    \includegraphics[width=0.48\textwidth]{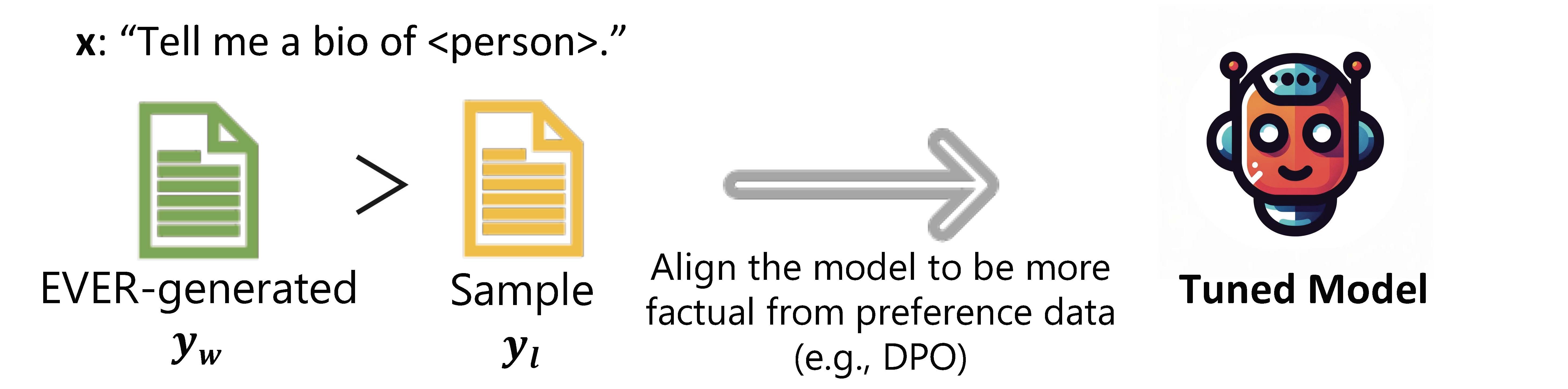}
    \caption{Illustration of using  \textsc{\textsc{EVER}}-generated data to construct preference data pairs, which are then used to further finetune the model to enhance the factuality of model.}
    \label{fig:teaser}
    \vspace{-1.5 em}
\end{figure}

In addition to directly rectify hallucination during the generation period, we extend the \textsc{EVER} framework to create better preference data to essentially enhance the factuality LLM by preference tuning. Here, as illustrated in Figure~\ref{fig:teaser}, the EVER-generated response $y_{ever}$ can be naturally served as preferred response and the non-rectified response is used as dispreferred response. Formally, the preference data is defined as $\mathcal{D}=\{x^{(i)}, y_w^{(i)}, y_l^{(i)}\}_{i=1}^N$, where $y_w^{(i)}=y_{ever}^{(i)}$ and $y_l^{(i)}$ represent preferred and dispreferred responses given an input prompt $x^{(i)}$. Such preference data are then used to perform preference tuning by using direct preference optimization (DPO)~\cite{rafailov2023direct}, which are detailed as follows:

Specifically, in large language model, we first define a language model policy $\pi_\theta$, which can produce the response $y$ with a conditional distribution $\pi_\theta(y\mid x)$. For each input prompt $x$ and response $y$, we define a reward function $r(x, y)$ to measure the generation quality of $y$. Our goal here is to maximize the average reward of outputs generated by the language model policy. Following a Bradley-Terry model~\citep{bradley1952rank}, DPO obtain each preference pair with the probability $p(y_w\succ y_l)$, which defined as:
\begin{equation}
    p(y_w\succ y_l)=\sigma(r(x, y_w)-r(x, y_l)),
\end{equation}
where $\sigma(\cdot)$ is defined as a sigmoid function. Then, DPO achieves the maximum average reward by optimizing the following classification loss over the preference data as:
\begin{equation}
\begin{split}
&\mathcal{L}(\pi_{\theta}, \mathcal{D}) = -\mathbb{E}_{(x,y_w,y_l) \sim \mathcal{D}} \\
&\left[ \log \sigma
\left(
\alpha \log \frac{\pi_\theta(y_w | x)}{\pi_{\text{ref}}(y_w | x)}
- \alpha \log \frac{\pi_\theta(y_l | x)}{\pi_{\text{ref}}(y_l | x)}
\right) \right],
\end{split}
\end{equation}
where $\pi_\text{ref}(y \mid x)$ is defined as the reference policy, typically referring to the result after performing supervised fine-tuning.

\section{Experiment of Prompting-based Hallucination Rectification}

In this section, we evaluate the performance of \textsc{\textsc{EVER}} on three tasks, short-form QA (Appendix~\ref{app:additonal}), biography generation and reasoning, aiming to answer the following questions: (1) Can \textsc{\textsc{EVER}} effectively address the challenges we've identified for RAG and post-hoc edit methods? (2) Can \textsc{\textsc{EVER}} effectively reduce hallucination of LLMs compared to other baselines across different tasks? (3) Can \textsc{\textsc{EVER}} effectively increase the trustworthiness of generated texts? In practice, we apply one of the following variant of \textsc{\textsc{EVER}} based on different application scenarios:
\vspace{-0.5em}
\begin{itemize}[leftmargin=*]
    \item \textbf{\textsc{\textsc{EVER}} (NRG+SQ)}: The first variant is a non-retrieval method that involves non-retrieval sentence generation (NRG) combined with a self-query (SQ) approach during the support check in the validation phase.
    \vspace{-0.5em}
    \item \textbf{\textsc{\textsc{EVER}} (NRG+ER)}: The second approach also employs a non-retrieval sentence generation approach, but it introduces evidence retrieval (ER) during the support check in the validation phase.
    \vspace{-1.5em}
    \item \textbf{\textsc{\textsc{EVER}} (RAG+ER)}: The third variant enhances sentence generation with retrieval-augmented methods (RAG) and includes evidence retrieval during support checking.
\end{itemize}

\subsection{Biography Generation Task}
In this task, the LLM is prompted to generate factual long-form biographies (bio), where LLM needs to ensure the accuracy of each atomic fact within the response. 
\subsubsection{Experimental Setup}
\textbf{Dataset and Evaluation Metric.} 
We utilize the bio benchmark with 183 examples as proposed by \cite{min2023factscore}, our model is prompted with "\textit{Tell me a bio of <entity>.}" to generate a biography for a given entity. To evaluate the effectiveness of our method, we employ the \textsc{FactScore} metric \cite{min2023factscore}. This metric leverages a retrieval-augmented language model ("ChatGPT + Retrieval"), for fact-checking the generated response, which has demonstrated that this metric aligns well with human evaluations. Furthermore, in line with other baseline settings, we retrieve evidence using Google Search.

\noindent \textbf{Evaluation Scenarios and Baselines.} We evaluate \textsc{EVER} in three scenarios: non-retrieval, retrieval-augmented rectification, and retrieval-augmented generation and rectification. Each scenario corresponds to a specific variant of \textsc{EVER}: \textsc{EVER (nrg+sq)}, \textsc{EVER (nrg+er)}, and \textsc{EVER (rag+er)}, respectively.

In each scenario, we employ different baselines for evaluation. First, in the non-retrieval scenario, we compare \textsc{EVER (nrg+sq)} with several models: 1) zero-shot generation models, including LLama 2 7B Chat, LLama 2 13B Chat \cite{touvron2023llama}, InstructGPT~\cite{ouyang2022training}, and GPT 3.5 Turbo; 2) a factuality-enhanced decoding method called Dola~\cite{chuang2023dola}; and 3) a chain of verification method called CoVE~\cite{dhuliawala2023chain}. Second, in the retrieval-based rectification scenario, we compare \textsc{EVER (nrg+er)} with RRAR\footnote{While the original paper uses Bing Search and GPT-3, we adapted the code to match our experimental setup with Serper Google Search API and our chosen LLMs.} \cite{gao2022attributed}. RRAR not only identifies attributions by using a search engine for outputs from various text generation models but also performs hallucination rectification. Third, for the RAG-like baselines, we compare \textsc{EVER (rag+er)} with vanilla RAG and Self-RAG~\cite{asai2023self}. These models are trained to retrieve, generate, and critique to enhance the LLM's output quality and factuality. Detailed descriptions of the baselines are discussed in Appendix~\ref{app:bio_basline}.

\subsubsection{Results and Analysis}
In Table~\ref{tab:bio_main}, we report the performance on the biography generation task. Specifically, we have the following observations: first, compared with non-retrieval based scenario with retrieval based scenario, we observe that external knowledge retrieval significantly enhances the factuality of text generation. This trend indicates that retrieval mechanisms enrich the inherent knowledge of large language models with up-to-date and specific information, thereby improving the content's accuracy. 

\begin{figure}[ht]
    \small
    \begin{center}
        \includegraphics[width=0.48\textwidth]{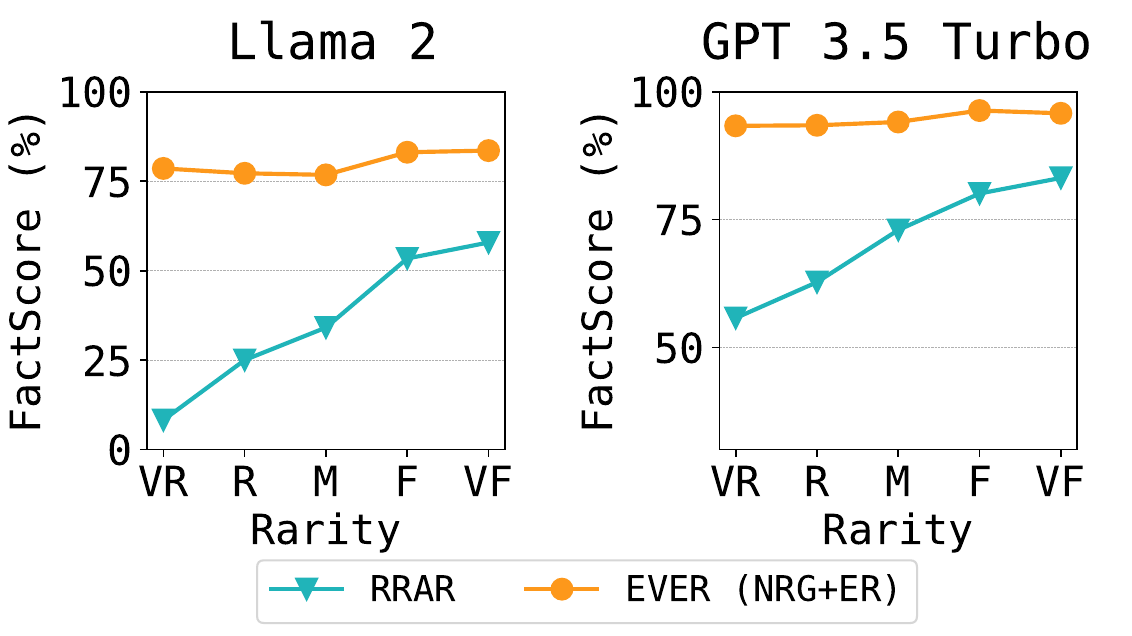}
        \caption{Comparison of our method and RRAR across examples with varying rarity distributions for the Llama 2 7B chat and GPT 3.5 Turbo models. "VR, R, M, F, VF" stands for "very rare", "rare", "medium", "frequent", and "very frequent", respectively.}
        \label{fig:rarity}
    \end{center}
    \vspace{-1.5em}
\end{figure}

Second, in comparison to other baselines of equivalent LLM scale, \textsc{EVER} exhibits superior performance in rectifying hallucinations across all scenarios, affirming the efficacy of its sentence-by-sentence generation, paired with real-time verification and rectification. In particular, when retrieval is not utilized, \textsc{EVER} outperforms the post-hoc verification and revision method CoVe when applied to the same pretrained Llama 65B model.
This effectiveness is further corroborated through a fine-grained comparison between \textsc{EVER} and RRAR. Here, we compare \textsc{EVER} and RRAR with respect to the rarity of the biography, as defined by the pageviews of their corresponding Wikipedia pages. The results in Figure~\ref{fig:rarity} illustrate that, unlike RRAR, which cannot reduce hallucinations for more rare subjects, the sentence-by-sentence evidence retrieval validation in \textsc{EVER} maintains stable factual precision across varying rarities.

Third, \textsc{EVER} could serve as a effective complementary method to the traditional retrieval-augmentation generation (RAG). Built upon traditional RAG, \textsc{EVER (rag+er)} demonstrates significant improvements over the conventional RAG approach. This demonstrates that \textsc{EVER} not only effectively retrieves relevant information but also adeptly incorporates and refines this information within the generated content.

\begin{table}[!ht]
    \caption{Results on the biography generation task. $^*$These numbers are from \citet{asai2023self}. $^\dagger$We obtain the results from \citet{dhuliawala2023chain}. $^\ddagger$The results are from \citet{min2023factscore}}.
    \vspace{-0.5em}
    \centering
    \small
    \resizebox{\linewidth}{!}{
    \setlength{\tabcolsep}{1mm}{
    \begin{tabular}{lcc}
        \toprule
        LM Scale & Method & \textsc{FactScore} (\%)  \\
        \midrule
        \multicolumn{3}{c}{\textbf{\textit{Non-Retrieval}}} \vspace{0.3em}\\
        InstructGPT  & Zero-Shot$^\ddagger$ &52.8\\
        
        -\ \ \ -\ \ \ -\ \ \ -\ \ \ -\ \ \ -\ \ \ -&-\ \ \ -\ \ \ -\ \ \ -\ \ \ -\ \ \ -\ \ \ -&-\ \ \ -\ \ \ -\ \ \ -\ \ \ -\ \ \ -\ \ \ - \\
        
        \multirow{3}{*}{Llama 2 7B Chat} & Zero-Shot &36.8 \\
                                        & Dola &36.8 \\
                                        &\textsc{EVER (nrg+sq)}& 46.7 \\
        
        -\ \ \ -\ \ \ -\ \ \ -\ \ \ -\ \ \ -\ \ \ -&-\ \ \ -\ \ \ -\ \ \ -\ \ \ -\ \ \ -\ \ \ -&-\ \ \ -\ \ \ -\ \ \ -\ \ \ -\ \ \ -\ \ \ - \\
        \multirow{3}{*}{Llama 2 13B Chat} & Zero-Shot &40.3 \\
                                        & Dola &40.1 \\
                                        &\textsc{EVER (nrg+sq)}& 47.5 \\
                                        
        -\ \ \ -\ \ \ -\ \ \ -\ \ \ -\ \ \ -\ \ \ -&-\ \ \ -\ \ \ -\ \ \ -\ \ \ -\ \ \ -\ \ \ -&-\ \ \ -\ \ \ -\ \ \ -\ \ \ -\ \ \ -\ \ \ -\\
        
        \multirow{3}{*}{Llama 1 65B} & Few-Shot$^\dagger$ & 55.9 \\
                                         & CoVe$^\dagger$ & 71.4  \\
                                        & \textsc{EVER (nrg+sq)} &72.9 \\
                                        
        -\ \ \ -\ \ \ -\ \ \ -\ \ \ -\ \ \ -\ \ \ -&-\ \ \ -\ \ \ -\ \ \ -\ \ \ -\ \ \ -\ \ \ -&-\ \ \ -\ \ \ -\ \ \ -\ \ \ -\ \ \ -\ \ \ -\\
        
        \multirow{2}{*}{GPT 3.5 Turbo} & Zero-Shot &71.8 \\
                                        &\textsc{EVER (nrg+sq)}& \textbf{75.2} \\

        \midrule

        \multicolumn{3}{c}{\textbf{\textit{Retrieval-Augmented Rectification}}} \\
        
        \multirow{2}{*}{Llama 2 7B Chat} & RRAR &37.8  \\
                                        & \textsc{EVER (nrg+er)} &76.9 \\        
        -\ \ \ -\ \ \ -\ \ \ -\ \ \ -\ \ \ -\ \ \ - & -\ \ \ -\ \ \ -\ \ \ -\ \ \ -\ \ \ -\ \ \ -&-\ \ \ -\ \ \ -\ \ \ -\ \ \ -\ \ \ -\ \ \ -\\

        \multirow{2}{*}{Llama 2 13B Chat} & RRAR &41.5  \\
                                        & \textsc{EVER (nrg+er)} &79.5 \\        
        -\ \ \ -\ \ \ -\ \ \ -\ \ \ -\ \ \ -\ \ \ -&-\ \ \ -\ \ \ -\ \ \ -\ \ \ -\ \ \ -\ \ \ -&-\ \ \ -\ \ \ -\ \ \ -\ \ \ -\ \ \ -\ \ \ -\\
        
        \multirow{2}{*}{GPT 3.5 Turbo} & RRAR & 74.3 \\
                                        & \textsc{EVER (nrg+er)} &\textbf{94.5} \\        
        \midrule

        \multicolumn{3}{c}{\textbf{\textit{Retrieval-Augmented Generation and Rectification}}} \vspace{0.3em}\\
        PerplexityAI  & RAG$^\ddagger$ & 71.2 \\ 
        \vspace{-1.5em} \\

        -\ \ \ -\ \ \ -\ \ \ -\ \ \ -\ \ \ -\ \ \ -&-\ \ \ -\ \ \ -\ \ \ -\ \ \ -\ \ \ -\ \ \ -&-\ \ \ -\ \ \ -\ \ \ -\ \ \ -\ \ \ -\ \ \ -\\
        
        \multirow{3}{*}{Llama 2 7B Chat} & RAG &79.4  \\
                                        & Self-RAG$^*$ & 81.2 \\
                                        & \textsc{EVER (rag+er)} & 86.4 \\        
        \vspace{-1.5em} \\
        -\ \ \ -\ \ \ -\ \ \ -\ \ \ -\ \ \ -\ \ \ -&-\ \ \ -\ \ \ -\ \ \ -\ \ \ -\ \ \ -\ \ \ -&-\ \ \ -\ \ \ -\ \ \ -\ \ \ -\ \ \ -\ \ \ -\\
        
        \multirow{3}{*}{Llama 2 13B Chat} & RAG$^*$ & 79.9  \\
                                     & Self-RAG$^*$ & 80.2 \\
                                        & \textsc{EVER (rag+er)} & 87.3 \\        
        - \ \ \ -\ \ \ -\ \ \ -\ \ \ -\ \ \ -\ \ \ -&-\ \ \ -\ \ \ -\ \ \ -\ \ \ -\ \ \ -\ \ \ -&-\ \ \ -\ \ \ -\ \ \ -\ \ \ -\ \ \ -\ \ \ -\\
        
        \multirow{2}{*}{GPT 3.5 Turbo} & RAG & 92.7  \\
                                    & \textsc{EVER (rag+er)} & \textbf{95.8}\\        
        \bottomrule
    \end{tabular}}}
    \label{tab:bio_main}
\end{table}

\subsection{Reasoning Task}
\label{sec:reason}
The final task we evaluate is the reasoning task, where the phenomenon of "hallucination snowballing" frequently arises~\cite{zhang2023language}. By leveraging the Chain-of-Thought (CoT) prompting method~\cite{wei2022chain}, we present multi-hop questions that required LLMs to construct an accurate and factually correct reasoning chain to provide the correct answers.

\subsubsection{Experimental Setup}
\textbf{Datasets and Experiment Settings.} 
In this task, follow~\citet{trivedi2022interleaving}, we use the test subset comprising 500 examples from the multi-hop question answering HotPotQA dataset \cite{yang2018hotpotqa}. We calculate the exact match (EM) and F1 score, following \cite{yang2018hotpotqa, gou2023critic}. For other experiment settings, we use the same setting as in the biography generation task.

\noindent \textbf{Baselines.} Similar to the biography task, we evaluate all variants of \textsc{EVER} in the reasoning task. For each variant, we employ differnt baselines for evaluation. First, in the non-retrieval scenario, we we compare \textsc{EVER (nrg+sq)} with Few-shot CoT~\cite{wei2022chain}. Second, in the retrieval-based rectification scenario, we compare \textsc{EVER (nrg+er)} with CRITIC~\cite{gou2023critic}. Thrid, we compare \textsc{EVER (rag+er)} with retrieval-based generation method IRCoT~\cite{trivedi2022interleaving}. See details about these baselines in Appendix~\ref{app:reason_baseline}.

\subsubsection{Results \& Analysis}
In Table~\ref{table:hotpotqa}, we report the results of \textsc{EVER} and other baselines on HotPotQA. According to the results, we demonstrate the superiority of \textsc{EVER} in improving the effectiveness of CoT prompting in reasoning tasks. Similar to the biography generation task, retrieval-based method significantly improves the performance compared with Few-Shot CoT. In addition, \textsc{EVER (nrg+er)} outperforms CRITIC, likely because CRITIC, while capable of verifying the final answers to multi-hop reasoning questions, corrects the reasoning chain as a whole rather than step-by-step. This approach cannot mitigate the "snowballing" issue throughout the steps. Moreover, by integrating retrieved knowledge prior to generation and incorporating a validation phase after generation, \textsc{EVER (rag+er)} outperforms the IRCoT method. This indicates the importance of both pre-generation retrieval and post-generation validation in enhancing the accuracy and reliability of CoT-based reasoning.

\begin{table}[ht]
\small
\centering
\caption{Results on the HotpotQA multi-hop reasoning dataset. $^*$The result is from \citet{gou2023critic}. }
\begin{tabular}{cccc}
\toprule
 \textbf{Retrieval} & \textbf{Method}  & \textbf{EM (\%)} & \textbf{F1 (\%)}  \\ \midrule
 \multirow{2}{*}{N/A}   & Few-Shot CoT   & 32.6 & 46.8    \\ 
    & \textbf{\textsc{EVER (nrg+sq)}} & \textbf{34.7} & \textbf{48.3} \\ \midrule
\multirow{2}{*}{Google} & RRAR & 34.5 & 46.7\\
 & CRITIC$^*$ & 40.3 & 52.9 \\ 
Dataset  & \textbf{\textsc{EVER (nrg+er)}} & \textbf{42.3}   & \textbf{58.1}    \\ \midrule
\multirow{2}{*}{Dataset}  & IRCoT    & 48.4 & 57.8     \\
 & \textbf{\textsc{EVER (rag+er)}}  & \textbf{51.4}     & \textbf{61.2}    \\ 
\bottomrule
\end{tabular}
\label{table:hotpotqa}
\end{table}

\begin{table*}[!ht]
\small
\centering
\caption{The three categories of extrinsic hallucination identified by ChatGPT based on human annotations, along with their respective percentages. We also list one representative validation question and the corresponding evidence, where the extracted concept for each validation question are marked in \colorbox{yellow}{yellow}.}
\begin{tabular}{@{}p{2.5cm}p{5.5cm}p{6cm}@{}} 
\toprule 
\multicolumn{1}{c}{\textbf{Category}} & \multicolumn{1}{c}{\textbf{Validation Question}} & \multicolumn{1}{c}{\textbf{Evidence}} \\
\midrule
Not mention (65\%) & Did notable achievements and impact in \colorbox{yellow}{Liga MX} earn Jorge Enríquez Garcías a debut for the Mexico national team? & \textit{...} Jorge Enríquez first played for the Mexico national team at the 2011 CONCACAF U-20 Championship \textit{...} \\
\addlinespace 
Need further inference (15\%)& Is Chris Johns one of the most dominant \colorbox{yellow}{featherweight champions} in boxing history? & \textit{...} Chris John was The Ring's \#8-ranked featherweight in the world (and \#10 pound-for-pound) \textit{...} \\
\addlinespace
Subjective (9\%)& Has Bobo Baldé left a \colorbox{yellow}{lasting impact} on the football world? & \textit{...} Dianbobo "Bobo" Baldé (born 5 October 1975) is a former professional footballer who played as a defender \textit{...} \\
\midrule
\multicolumn{3}{c}{Misclassified examples (11\%)} \\
\bottomrule 
\end{tabular}
\label{tab:extrinsic}
\vspace{-1em}
\end{table*}

\subsection{Analysis}
\textbf{Extrinsic Hallucination Analysis.} In the biography generation task, we conduct a human annotation analysis of the 300 instances that are classified as "Not Enough Info" (NEI). Here, we define three distinct categories of extrinsic hallucination, as showed in table \ref{tab:extrinsic}. The most prevalent cases, found in 65\% of cases, is that the evidence provided does not directly contain relevant information to support or contradict. The second most common error of the generated text, accounting for 15\% of the instances, is that while the evidence is relevant, it requires additional inference. Also, 9\% of cases involve subjective, opinion-based or interpretative content that is hard to classify objectively. Finally, our findings reveal that \textsc{EVER} incorrectly categorizes 11\% of examples as "Not Enough Info" (NEI), despite these instances actually being supportive or contradictory. Nevertheless, the high accuracy of NEI-classified examples demonstrates both \textsc{EVER}'s strong performance and the practicality of user warnings, cautioning against potential lack of factuality.

\noindent \textbf{Efficiency Analysis.} Although the proposed active concept-level validation and rectification in \textsc{EVER} incurs time overheads, these overheads are typical in similar retrieval-based baselines. As Table~\ref{tab:runtime} illustrates, all three \textsc{EVER} variants demonstrate runtime comparable to those of other methods in biography generation and multi-hop reasoning. The efficiency of \textsc{EVER} results from the simplification of tasks into shorter, few-shot, or zero-shot prompts and the parallel validation of extracted concepts.
\begin{table}[b]
\small
\centering
\caption{Average runtime (s) comparison across different methods on the two datasets for the GPT 3.5 Turbo model. $^*$For CRITIC, involving up to three iterations, we calculate the average runtime.}
\begin{tabular}{lcc}
\toprule
\textbf{Method} & \textbf{Biography} & \textbf{HotpotQA} \\
\midrule
RRAR            & 210.5                  & -                      \\
IRCoT           & -                      & 67.2                   \\
CRITIC$^*$          & -                      & 83.8                   \\\midrule
\textsc{EVER (nrg+sq)}    & 195.7                 & 73.6                   \\
\textsc{EVER (nrg+er)}    & 141.8                  & 86.9                   \\
\textsc{EVER (rag+er)}    & 115.4                  & 62.8                   \\
\bottomrule
\end{tabular}
\label{tab:runtime}
\vspace{-1em}
\end{table}

\section{Experiment of Enhancing Factuality with Preference Tuning}
In this section, we study the performance of finetuning language models by using the EVER-generated preference data pair to reduce hallucination.

\subsection{Experimental Setup}
\noindent \textbf{Datasets and Experiment Settings}. We adopt the aforementioned biography generation task and use the same 183 human entities as the test set to evaluate the fine-tuning result. Follow~\citet{tian2023fine}, we use the EVER-generated data as the preferred sample and other 20 randomly zero-shot generations for each human entity as the dispreferred sample. In total, we have 10,000 training preference pairs.

\noindent \textbf{Baselines.} We compare several methods, including the vanilla approach, which uses the SFT model output. Another baseline is FactTune-FS~\cite{tian2023fine}, which samples 10 generations and runs DPO on ${10 \choose 2}$ pairs with using FactScore to select the better one in each pair. Both of these two approaches compare with \textsc{EVER-Pref (NRG+SQ)}. In addition, we use the vanilla RAG-generated data (RAG-\textsc{Pref}) as the chosen text as a baseline to create preference data pairs, which is then compared with the retrieval-based version of \textsc{EVER-Pref}.

\subsubsection{Results \& Analysis.}
As shown in Table~\ref{tab:ft_main}, fine-tuning the Llama-2-7B-chat model on the biography generation task has yielded insights into reducing hallucination in language models. Initially, finetuning using the generated data by retrieval-free and self-query versions of \textsc{EVER} demonstrates a reduction in hallucinations, as evidenced by the improvement in FactScore from the Vanilla baseline of 36.8\% to 47.3\% with \textsc{EVER-Pref (nrg+sq)}. This indicates that fine-tuning with retrieval-free methods enhances the factual accuracy of language models.

Further advancements are observed when fine-tuning incorporated text generated through retrieval mechanisms. Specifically, the utilization of more factual data by retrieval during fine-tuning, particularly with \textsc{EVER-Pref (nrg+er)} and \textsc{EVER-Pref (rag+er)}, increases the performance even further, achieving FactScores of 52.8\% and 53.9\% respectively. These results underscore the potential of fine-tuning language models with factually enriched datasets to mitigate hallucinations.

\begin{table}[ht]
    \caption{Results of finetuning the Llama-2-7B-chat model on the biography generation task.}
    \vspace{-0.5em}
    \centering
    \small
    \begin{tabular}{lc}
        \toprule
        Method & \textsc{FactScore} (\%)  \\
        \midrule


          Vanilla &36.8 \\
          FactTune-FS & 45.4 \\
          \textsc{EVER-Pref (nrg+sq)}& \textbf{47.3} 
         \\\midrule
          RAG-\textsc{Pref} & 50.2  \\
          \textsc{EVER-Pref (nrg+er)} &52.8 \\ 
          \textsc{EVER-Pref (rag+er)} &\textbf{53.9} \\ 
        \bottomrule
    \end{tabular}
    \label{tab:ft_main}
    \vspace{-1.5em}
\end{table}

\section{Related Work}

\noindent \textbf{Hallucination Detection.} Detecting hallucinations in LLMs is crucial for ensuring the reliability of generated content. To detect LLM hallucination, the first line of methods analyze the probability of tokens \cite{mielke2022reducing, kadavath2022language, varshney2023stitch}. Another line of methods leverage the inconsistency between multiple generated examples, including NLI-based approaches~\cite{elaraby2023halo,manakul2023selfcheckgpt} and QA-based methods~\cite{manakul2023selfcheckgpt,agrawal2023language}. In addition, \citet{cohen2023lm} introduced a method in which one LM acts as an examiner, repeatedly cross-examining the outputs of the other LM to test their consistency. 

\noindent \textbf{Hallucination Mitigation.}
A number of approaches have been developed to mitigate hallucination in LLMs. 
One line of work focuses on manipulating the model via decoding strategies~\cite{chuang2023dola, shi2023trusting, li2022contrastive, li2023inference} or preference fine-tuning~\cite{tian2023ft}. Another line of work uses post-hoc edit methods, which can be further divided into those involving retrieval \cite{peng2023check, menick2022teaching, gao2022attributed, chern2023factool, yu2023improving, varshney2023stitch} and non-retrieval based strategies \cite{dhuliawala2023chain,zhou2023analyzing}. RAG is another approach to improve factuality by integrating external knowledge during the generation process \cite{lewis2020retrieval, jiang2023active, asai2023self}. Yet, non-retrieval-based methods lack of updated information, RAG lacks of robustness to irrelevant and useless context, and post-hoc editing methods may not address the snowballing issue of hallucinations. Our proposed method, with step-by-step verification and rectification, effectively mitigates these challenges in prior work. In addition, we show that our proposed method \textsc{EVER} can be utilized to create better preference data to further finetune a LLM to be enhance its factuality.

\noindent \textbf{Reasoning Improvement.} Several studies aim to enhance LLMs' performance in reasoning tasks. One line of works uses prompting strategies~\cite{wei2022chain, zhou2022least,  kojima2022large, wang2022self} to divide a difficult task into simpler ones and/or utilizes external tools to aid LLMs~\cite{yao2022react, schick2023toolformer, gao2023pal, yang2022generating}, both of which are solving problems sequentially without checking the correctness of generation. Also, \citet{gou2023critic, zhao2023verify} involves post-generation verification. However, these works only focus on reasoning tasks, making it difficult to generalize to non-reasoning tasks. Additionally, they don't improve the trustworthiness of generated texts. We take these challenges into consideration, and \textsc{EVER} utilizes general-purpose verification and rectification strategies that are suitable for various tasks. Furthermore, the user warning further enhances the trustworthiness of generated texts.

\section{Conclusion}
In this paper, we introduce the \textsc{EVER}, aiming to mitigate hallucination in LLMs. \textsc{EVER} effectively addresses both intrinsic and extrinsic hallucinations while also reducing the propagation of errors that may occur in sequential text generation. Our empirical results demonstrate that \textsc{EVER} significantly reduces hallucination in various tasks, including short-form QA, long-form biography generation, and reasoning. Moreover, \textsc{EVER} is able to generate better preference data pair to further finetune the model to reduce hallucination.

\section*{Limitation}
\label{app:limit}
This study acknowledges limitation in the \textsc{EVER} framework. Unlike conventional fact-checking process, which involves considering the information beyond the evidence (e.g., claimant, claim date, source, etc.) to check the factual accuracy,  our focus is solely on enhancing text attribution to reduce hallucinations. This only require an reference (which might be incorrect) that could support a fact.

\section*{Acknowledgement}
We thank the Center for AI Safety and Google Cloud Research Credits program for supporting our computing needs.

\bibliographystyle{acl_natbib}
\bibliography{anthology,custom}

\appendix

\label{sec:appendix}

\section{Additional Experiment on Short-form QA Task}
\label{app:additonal}
Honesty-tuned LLMs may exhibit over-conservatism due to an imbalanced trade-off between helpfulness and honesty \cite{ouyang2022training}. In this short-form QA task, we evaluate \textsc{\textsc{EVER}}'s ability to strike a better balance in this trade-off. Employing open-domain questions, \textsc{\textsc{EVER}} is designed to either abstain from answering or to modify answers depending on the context, aiming for generating more trustworthy text.

\subsection{Experimental Setup}
\noindent \textbf{Dataset.} In this task, we use two short-form QA datasets, including TriviaQA-unfiltered \cite{joshi2017triviaqa} and ALCE-Qampari QA~\cite{gao2023enabling}. For TriviaQA, we assume there is only one correct answer for each question. Since the test set of TriviaQA is not publicly available, we use the same test split from validation set as~\citet{min-etal-2019-discrete, asai2023self}. 

\noindent \textbf{Evaluation Metric.} Following \citet{schick2023toolformer}, we evaluate performance based on whether gold answers are included in the model generations, rather than strictly requiring an exact string match. We report accuracy on the answered examples as $N_c / (N_{all} - N_{rej})$, and the percentage of trustworthy examples as $(N_c + N_{rej}) / N_{all}$, where $N_c$, $N_{rej}$, and $N_{all}$ represent the number of correct examples, abstention examples, and all examples, respectively. For Qampari QA, where the gold answer is a list of answers, we follow \citet{gao2023enabling, schick2023toolformer} in evaluating performance using the \textit{recall@5} metric. Here, we consider recall to be 100\% if the prediction includes at least 5 correct answers. Additionally, we assess the \textit{precision} of the model's prediction by checking for an exact string match with the gold answer list.

\noindent \textbf{Baselines.}
We evaluate \textsc{\textsc{EVER}} against two categories of baseline approaches: (1) zero-shot generation and vanilla retrieval-augmented generation, and (2) improvements to the baselines in category (1) by prompting LLMs to abstain from uncertain examples. In the zero-shot and RAG approaches with abstention prompting, LLMs respond with \textit{"Sorry, I don't know"} when unsure or when retrieved evidence is insufficient to answer, respectively. See detailed discussions in Appendix~\ref{app:baseline}.

\noindent \textbf{Experiment Settings.}
We employ two methods to retrieve relevant evidence: Google and the dataset. For each question, we retrieve the top 5 relevant documents from the provided dataset. When using Google, we retrieve a total of 10 relevant documents by querying both the question and the concatenation of the question and answer strings.
\begin{table*}[t]
\small
\centering
\caption{The results of GPT 3.5 turbo on the Trivia QA and Qampari QA datasets.}
    \resizebox{\linewidth}{!}{
\setlength{\tabcolsep}{1.6mm}{
\begin{tabular}{lc|ccc|ccc}
\toprule
\multirow{2}{*}{\textbf{Method}} & \multirow{2}{*}{\textbf{Retrieval}} & \multicolumn{3}{c|}{\textbf{Trivia QA}} & \multicolumn{3}{c}{\textbf{Qampari QA}} \\
 &  & \textbf{Accuracy} & \textbf{\%Trustful} & \textbf{\%Abstention} & \textbf{Recall@5} & \textbf{Precision} & \textbf{\%Abstention}\\
 \midrule
Zero-shot & \multirow{2}{*}{N/A} &76.7 & 76.7 & - & 11.6 & 16.8 & - \\
Zero-shot+prompting & & 80.4 & 79.0 & 11.7 & 11.4 & 33.5 & 46.0 \\
\textbf{\textsc{EVER (nrg+er)}} & Dataset &83.4 & 82.8 & 3.0 & 11.8& 26.6 & 9.0 \\
\midrule
RAG & \multirow{3}{*}{Dataset} &71.3 & 71.3 & - & 22.8 & 35.2 & - \\
RAG+prompting &  & 79.2 & 80.3 & 14.7 & 22.7 & 38.9 & 29.5 \\
\textbf{\textsc{EVER (rag+er)}} &  & 82.3 & 86.8 & 5.3 & \textbf{23.3} & \textbf{39.2} & 1.0 \\
\midrule
RAG  & \multirow{3}{*}{Google} &79.0 & 79.0 & - & - & - & - \\
RAG+prompting &  &81.3 & 82.0 & 10.0 & - & - & - \\
\textbf{\textsc{EVER (rag+er)}} &  & \textbf{84.9} & \textbf{87.7} & 4.0 & - & - & - \\\bottomrule
\end{tabular}}}
\label{table:shortformQA}
\vspace{-1em}
\end{table*}
\subsection{Results and Analysis}
Table~\ref{table:shortformQA} reveals that traditional abstention prompting-based methods, as highlighted in~\citet{ouyang2022training}, tend to exhibit over-conservatism by refusing to answer a significant number of questions across datasets. In contrast, our \textsc{EVER} method stands out for its inclination to provide correct answers rather than abstaining, significantly enhancing the helpfulness of the generated text. Additionally, \textsc{EVER} outperforms other baselines in trustworthiness, as evidenced by its higher trustful rate in Trivia QA. Furthermore, \textsc{EVER} demonstrates strong performance in producing higher correctness/factuality, showing higher accuracy, precision and recall compared to other baselines. Finally, \textsc{EVER} with evidence retrieval can also address the limitations of RAG. In the Trivia QA dataset, RAG performs even worse compared with zero-shot generation when using the top-5 retrieved documents from the provided dataset as context, often due to the inclusion of irrelevant or misleading text. However, this issue can be effectively resolved by employing \textsc{EVER}. In summary, \textsc{EVER} effectively balances the trade-off between helpfulness and honesty, ensuring that the text it generates is both informative and reliable.

\section{Multi-Round Rectification}
We evaluate the effects of allowing multi-round rectification for GPT 3.5 Turbo model. The results in Table~\ref{tab:multi-round} shows that in general one round of rectification is sufficient for both tasks. Additional rounds of rectification yield negligible improvements in performance.
\label{app:multi-round}
\begin{table}[ht]
\small
    \caption{The results of multi-round rectification of \textsc{EVER (nrg+er)} on the biography generation and reasoning tasks for GPT 3.5 Turbo. }
    \centering
    \begin{tabular}{cccc}
        \toprule 
        \# Rounds &  \textsc{\textsc{FactScore}} (\%)  & EM (\%) & F1 (\%)\\
        \midrule
         1 & 94.5 & 42.3 & 58.1 \\ 
         2 & 94.7 & 43.5 & 57.8\\ 
         3 & 95.2 & 43.1 & 59.4\\ 
         4 & 93.8 & 42.6 & 58.3\\
         \bottomrule
    \end{tabular}
    \label{tab:multi-round}
\end{table}

\section{Short-form QA Baselines}
\label{app:baseline}
Zero-shot involves generating texts solely based on the provided prompt without any additional contextual information. Retrieval Augmented Generation (RAG) incorporates an external knowledge in the prompt to enhance the generation process. RAG has two sources: relevant documents provided in the original datasets and relevant documents obtained through Google Search. For prompting, we employ prompting engineering to increase the trustworthiness of generated text by instructing the model to respond with \textit{"I don't know"}  if there is no answer within the context. The model's response \textit{"I don't know"} is considered an abstention. For the methods of zero-shot, zero-shot+prompting, RAG, and RAG+prompting, as well as different datasets, we use different prompts, which are listed in Table~\ref{tab:qampari_prompt} and Table~\ref{tab:trivia_prompt}.

\section{Biography Generation Baselines}
\label{app:bio_basline}
\begin{itemize}
    \item \textbf{Dola:} This decoding method leverages the observed phenomenon that certain transformer layers within LLMs tend to localize factual knowledge. It computes the distribution for the next token by comparing the logit discrepancies when mapped to the vocabulary from later versus earlier layers.
    \item \textbf{CoVe:} In this non-retrieval-based pipeline, a LM sequentially drafts a response, devises fact-checking queries, independently answers them to avoid bias, and finally produces a verified response.
    \item \textbf{RRAR:} This approach automatically attributes the generated text from any model and subsequently refines the output to rectify any unsupported content, striving to maintain the integrity of the initial output.
    
    \item \textbf{Self-RAG:} This method improves an LM's output quality and accuracy by incorporating retrieval and self-reflection. It trains an LM to fetch relevant passages as needed and to introspect on both the passages and its own generated content with "reflection tokens." These tokens allow for controlled inference, making the LM flexible for various tasks.
\end{itemize}

\section{Reasoning Baselines}
\label{app:reason_baseline}
\begin{itemize}
    \item \textbf{CRITIC:} This method enables LLMs to self-validate and iteratively refine their outputs, mimicking human revision processes. It begins with an initial output and utilizes tools to assess and enhance text quality based on the feedback received.
    \item \textbf{IRCoT:} This work integrates retrieval into the Chain of Thought process, using each step to direct retrieval and leveraging the gathered information to bolster the reasoning chain.
\end{itemize}

\section{Prompt Templates}
\begin{table*}[tp]
\small
\centering
    \caption{The prompts used to generate answers for the QampariQA dataset.}

    \begin{tabular}{p{\linewidth}}
    \toprule

    \textbf{Zero-shot}\\
    Provide a list of accurate answers for the given question using only the provided context (some of which might be irrelevant). Separate answers by semicolons. For questions that have more than 5 answers, write at least 5 answers.\\
    \textbf{Question:} ...\\
    \textbf{Answer:}\\
    \\
    \textbf{Zero-shot+prompting}\\
    Provide a list of accurate answers for the given question using only the provided context (some of which might be irrelevant). Separate answers by semicolons. For questions that have more than 5 answers, write at least 5 answers. If there is no answer in the context, reply ``sorry I don't know''.\\
    \textbf{Question:} ...\\
    \textbf{Answer:}\\
    \\
    \textbf{RAG}\\
    \textbf{Context:} ...\\
    Provide a list of accurate answers for the given question using only the provided context (some of which might be irrelevant). Separate answers by semicolons. For questions that have more than 5 answers, write at least 5 answers.\\
    \textbf{Question:} ...\\
    \textbf{Answer:}\\
    \\
    \textbf{RAG+prompting}\\
    \textbf{Context:} ...\\
    Provide a list of accurate answers for the given question using only the provided context (some of which might be irrelevant). Separate answers by semicolons. For questions that have more than 5 answers, write at least 5 answers. If there is no answer in the context, reply ``sorry I don't know''.\\
    \textbf{Question:} ...\\
    \textbf{Answer:}\\
    \\

    \bottomrule

    \end{tabular}
    \label{tab:qampari_prompt}
\end{table*}

\begin{table*}[tp]
\small
    \caption{The prompts used to generate answers for the TriviaQA dataset.}

\centering

    \begin{tabular}{p{\linewidth}}
    \toprule

    \textbf{Zero-shot}\\
    Answer the following question.\\
    \textbf{Question:} ...\\
    \textbf{Answer:}\\
    \\
    \textbf{Zero-shot+prompting}\\
    Answer the following question based on the context. If there is no answer in the context, reply ``sorry I don't know''.\\
    \textbf{Question:} ...\\
    \textbf{Answer:}\\
    \\
    \textbf{RAG}\\
    \textbf{Context:} ...\\
    Answer the following question based on the context.\\
    \textbf{Question:} ...\\
    \textbf{Answer:}\\
    \\
    \textbf{RAG+prompting}\\
    \textbf{Context:} ...\\
    Answer the following question based on the context. If there is no answer in the context, reply ``sorry I don't know''.\\
    \textbf{Question:} ...\\
    \textbf{Answer:}\\
    \\

    \bottomrule

    \end{tabular}
    \label{tab:trivia_prompt}
\end{table*}

\begin{table*}[ht]
\small
\centering
    \caption{The prompts used to extract concepts.}

    \begin{tabular}{p{0.9\linewidth}}
    \toprule
        \textbf{Instruction:} Identify all objective factual concepts from the following sentence. Exclude the main subject and any subjective terms. Include all numerical details (such as times, quantities, etc.). Present your findings in a list separated by semicolons. \\
        \textbf{Sentence:} Claude Monet (14 November 1840 – 26 December 1926) was a French painter born in Rue Laffitte, Paris, France, who along with his companions Auguste Renoir, Edgar Degas and Pierre-Auguste Renoir, is often referred to as the founder of Impressionism.\\
        \textbf{Answer:} 14 November 1840; 26 December 1926; Rue Laffitte, Paris, France; French; painter; Auguste Renoir; Edgar Degas; Pierre-Auguste Renoir; founder of Impressionism\\
        \\
        \textbf{Instruction:} Identify all objective factual concepts from the following sentence. Exclude the main subject and any subjective terms. Include all numerical details (such as times, quantities, etc.). Present your findings in a list separated by semicolons. \\
        \textbf{Sentence:} Lee Min-ho has also won several awards for his outstanding performances in popular films like "Gangnam Blues" and "Bounty Hunters."\\
        \textbf{Answer:} awards; popular films; Gangnam Blues; Bounty Hunters\\
        \\
        \textbf{Instruction:} Identify all objective factual concepts from the following sentence. Exclude the main subject and any subjective terms. Include all numerical details (such as times, quantities, etc.). Present your findings in a list separated by semicolons. \\
        \textbf{Sentence:} Pablo Escobar, often referred to as "El Patrón," was a Colombian drug lord and the leader of the Medellín Cartel, dominating the cocaine trade during the 1970s and 1980s.\\
        \textbf{Answer:} El Patrón; Colombian; drug lord; Medellín Cartel; cocaine trade; 1970s; 1980s\\
        \\
        \textbf{Instruction:} Identify all objective factual concepts from the following sentence. Exclude the main subject and any subjective terms. Include all numerical details (such as times, quantities, etc.). Present your findings in a list separated by semicolons. \\
        \textbf{Sentence:} Meryl Streep earned widespread acclaim for her performances in films like "The Iron Lady," "Doubt," and "Julie \& Julia."\\
        \textbf{Answer:} The Iron Lady; Doubt; Julie \& Julia\\
        \\
        \textbf{Instruction:} Identify all objective factual concepts from the following sentence. Exclude the main subject and any subjective terms. Include all numerical details (such as times, quantities, etc.). Present your findings in a list separated by semicolons. \\
        \textbf{Sentence:} \{sentence\} \\
        \textbf{Answer:} 
        \\
        \bottomrule

    \end{tabular}
\end{table*}

\begin{table*}[tp]
\small
\centering
    \caption{The prompts used to generate validation questions for smaller models, such as Llama 2 7B/13B Chat. For GPT-3.5, we use zero-shot with the same instruction.}

    \begin{tabular}{p{\linewidth}}
    \toprule
    \textbf{Sentence:} Leonardo da Vincian, an Italian polymath of the High Renaissance who was active as a painter, draughtsman, engineer, scientist, theorist, sculptor, and architect, was born in Vinci, Italy, on 15 April 1452. \\
    For the above sentence about "Leonardo da Vinci", generate a yes/no question WITHOUT any pronouns about the entity of "15 April 1452". The question MUST contain the entity. \\
    \textbf{Question:} Was Leonardo da Vinci born on 15 April 1452? \\
    \\
    \textbf{Sentence:} Wolfgang Amadeus Mozart, during his brief lifetime, composed more than 600 works, many of which are acknowledged as the pinnacles of symphonic, concertante, chamber, operatic, and choral music.\\
    For the above sentence about "Wolfgang Amadeus Mozart", generate a yes/no question WITHOUT any pronouns about the entity of "more than 600 works". The question MUST contain the entity. \\
    Question: Did Wolfgang Amadeus Mozart compose more than 600 works during his lifetime? \\
    \\
    \textbf{Sentence: }Frida Kahlo, a renowned Mexican artist, is best known for her self-portraits and works like "The wounded deer" and "The Two Fridas".\\
    For the above sentence about "Frida Kahlo", generate a yes/no question WITHOUT any pronouns about the entity of "The Two Fridas". The question MUST contain the entity.\\
    \textbf{Question:} Did Frida Kahlo create "The Two Fridas"?\\
    \\
    \textbf{Sentence: }\{sentence\}\\
    For the above sentence about "\{topic\}", generate a yes/no question WITHOUT any pronouns about the entity of "\{topic\}". The question MUST contain the entity.\\
    \textbf{Question:}\\
    \bottomrule

    \end{tabular}
\end{table*}

\begin{table*}[!ht]
\small
\centering
\caption{The prompts used to do support checking with evidence retrieval.}
\begin{tabularx}{\linewidth}{Xl}
    \toprule \\
Based on the evidence, answer the following question by selecting one of these options: True, False, or Not Enough Information. YOU MUST PROVIDE THE REASONING FIRST BEFORE MAKING A DECISION.\\
\textbf{Evidence: }Jane Austen - BritishLiteratureArchive.org: Jane Austen (16 December 1775 – 18 July 1817) was an English novelist known for her novels that critique the British landed gentry of the 18th century.\\
\textbf{Question: }Was Jane Austen an English novelist?\\
\textbf{Answer: }The evidence presents Austen as an English novelist. The claim is consistent with this information. Therefore, the decision is True.\\
\\
Based on the evidence, answer the following question by selecting one of these options: True, False, or Not Enough Information. YOU MUST PROVIDE THE REASONING FIRST BEFORE MAKING A DECISION.\\
\textbf{Evidence:} Ada Lovelace - WomenInTechHistory.com: Ada Lovelace (10 December 1815 – 27 November 1852) was an English mathematician and writer, chiefly known for her work on Charles Babbage's proposed mechanical general-purpose computer, the Analytical Engine.\\
\textbf{Question:} Is Ada Lovelace regarded as the first computer programmer?\\
\textbf{Answer: }The evidence describes Ada's significant work on the Analytical Engine, a proposed mechanical computer by Charles Babbage. However, it doesn't explicitly state that she is considered the first computer programmer. Therefore, the decision is Not Enough Information.\\
\\
Based on the evidence, answer the following question by selecting one of these options: True, False, or Not Enough Information. YOU MUST PROVIDE THE REASONING FIRST BEFORE MAKING A DECISION.\\
\textbf{Evidence:} Leonardo da Vinci - RenaissanceMasters.org: Leonardo da Vinci (15 April 1452 – 2 May 1519) was an Italian polymath of the Renaissance era, known for his works in painting, science, mathematics, and various other fields.\\
\textbf{Question:} Was Leonardo da Vinci a 17th-century composer known for his operas?\\
\textbf{Answer:} The evidence introduces da Vinci as an Italian polymath from the Renaissance era, acclaimed for his contributions in painting, science, and other areas. The claim erroneously describes him as a 17th-century composer, which doesn't align with the known facts. Therefore, the decision is False.\\
\\
Based on the evidence, answer the following question by selecting one of these options: True, False, or Not Enough Information. Multiple sources of evidence are presented, each separated by a semicolon. YOU MUST PROVIDE THE REASONING FIRST BEFORE MAKING A DECISION. \\
\textbf{Evidence:} \{evidence\}\\
\textbf{Question:} \{validation question\}\\
\textbf{Answer:} \\
\bottomrule 
\end{tabularx}
\end{table*}

\begin{table*}[!ht]
\small
\centering
\caption{The prompts used to do support checking with self-query. We use an \textit{"According to"} prompting strategy to better recall memorized facts in LMs~\cite{weller2023according}.}

\begin{tabularx}{\linewidth}{Xl}
    \toprule \\
Answer the following question by selecting one of these options: True, False, or Not Enough Information. YOU MUST PROVIDE THE REASONING FIRST BEFORE MAKING A DECISION.\\
\textbf{Question:} Was Jane Austen an English novelist?\\
\textbf{Answer:} According to Wikipedia, Jane Austen (1775-1817) was an English novelist who is best known for her six major novels, including "Pride and Prejudice," "Sense and Sensibility," and "Emma." Therefore, the decision is True.\\
\textbf{Question:} Was Jane Austen an English novelist?\\
\textbf{Answer:} According to Wikipedia, Jane Austen (1775-1817) was an English novelist who is best known for her six major novels, including "Pride and Prejudice," "Sense and Sensibility," and "Emma." Therefore, the decision is True.\\

Answer the following question by selecting one of these options: True, False, or Not Enough Information. YOU MUST PROVIDE THE REASONING FIRST BEFORE MAKING A DECISION.\\

\textbf{Question:} Was Jane Austen an English novelist?\\
\textbf{Answer:} According to Wikipedia, Jane Austen (1775-1817) was an English novelist who is best known for her six major novels, including "Pride and Prejudice," "Sense and Sensibility," and "Emma." Therefore, the decision is True.\\

Answer the following question by selecting one of these options: True, False, or Not Enough Information. YOU MUST PROVIDE THE REASONING FIRST BEFORE MAKING A DECISION.\\
\textbf{Question:} Is Ada Lovelace regarded as the first computer programmer?\\
\textbf{Answer:} According to Wikipedia, Ada Lovelace (1815-1852) was an English mathematician and writer, known for her work on Charles Babbage's early mechanical general-purpose computer, the Analytical Engine. No further information about her high school love is mentioned on Wikipedia. Therefore, the decision is Not Enough Information.\\
\\
Answer the following question by selecting one of these options: True, False, or Not Enough Information. YOU MUST PROVIDE THE REASONING FIRST BEFORE MAKING A DECISION.\\

\textbf{Question:} Was Leonardo da Vinci a 17th-century composer known for his operas?\\
\textbf{Answer:} According to Wikipedia, Leonardo da Vinci as an Italian polymath from the Renaissance era, acclaimed for his contributions in painting, science, and other areas. The claim erroneously describes him as a 17th-century composer, which doesn't align with the known facts. Therefore, the decision is False.\\

Answer the following question by selecting one of these options: True, False, or Not Enough Information. YOU MUST PROVIDE THE REASONING FIRST BEFORE MAKING A DECISION.\\
\textbf{Question:} \{validation question\} \\
\textbf{Answer:} According to Wikipedia, \\

\bottomrule 
\end{tabularx}
\end{table*}

\end{document}